\pdfoutput=1

\documentclass[11pt]{article}

\usepackage[final]{acl}

\usepackage{times}
\usepackage{latexsym}

\usepackage[T1]{fontenc}

\usepackage[utf8]{inputenc}

\usepackage{microtype}

\usepackage{inconsolata}

\usepackage{graphicx}

\usepackage[linesnumbered,ruled]{algorithm2e}
\usepackage{multirow}
\usepackage{pifont}
\usepackage{colortbl}
\usepackage{xspace}

\newcommand{\eg}{\textit{e.g.}\xspace}
\newcommand{\ie}{\textit{i.e.}\xspace}

\usepackage{enumitem}
\AtEndPreamble{
    \usepackage[capitalize]{cleveref}
    \crefname{section}{Sec.}{Secs.}
    \Crefname{section}{Section}{Sections}
    \crefname{table}{Tab.}{Tabs.}
    \Crefname{table}{Table}{Tables}
}
\usepackage{amsmath}
\usepackage{booktabs}
\usepackage{subcaption}
\usepackage{pifont}
\usepackage{url}
\title{QAVA: Query-Agnostic Visual Attack to Large Vision-Language Models}

\author{
 \textbf{Yudong Zhang\textsuperscript{1,2}},
 \textbf{Ruobing Xie\textsuperscript{2,\ding{41}}},
 \textbf{Jiansheng Chen\textsuperscript{3,\ding{41}}},
 \textbf{Xingwu Sun\textsuperscript{2,4}},
\\
 \textbf{Zhanhui Kang\textsuperscript{2}},
 \textbf{Yu Wang\textsuperscript{1,5,\ding{41}}}
\\
 \textsuperscript{1}Department of Electronic Engineering, Tsinghua University,\\
 \textsuperscript{2}Machine Learning Platform Department, Tencent,\\
 \textsuperscript{3}School of Computer and Communication Engineering, University of Science and \\ Technology Beijing,
 \textsuperscript{4}Faculty of Science and Technology, University of Macau,\\
 \textsuperscript{5}State Key Laboratory of Space Network and Communications, Tsinghua University
\\
\small{
 zhangyd16@mails.tsinghua.edu.cn, xrbsnowing@163.com, jschen@ustb.edu.cn, sunxingwu01@gmail.com, 
}
\\
\small{
kegokang@tencent.com, yu-wang@mail.tsinghua.edu.cn. (\ding{41}: Corresponding authors)
}
}

\begin{document}
\maketitle
\begin{abstract}
  In typical multimodal tasks, such as Visual Question Answering (VQA), adversarial attacks targeting a specific image and question can lead large vision-language models (LVLMs) to provide incorrect answers. However, it is common for a single image to be associated with multiple questions, and LVLMs may still answer other questions correctly even for an adversarial image attacked by a specific question. To address this, we introduce the query-agnostic visual attack (QAVA), which aims to create robust adversarial examples that generate incorrect responses to unspecified and unknown questions. Compared to traditional adversarial attacks focused on specific images and questions, QAVA significantly enhances the effectiveness and efficiency of attacks on images when the question is unknown, achieving performance comparable to attacks on known target questions. Our research broadens the scope of visual adversarial attacks on LVLMs in practical settings, uncovering previously overlooked vulnerabilities, particularly in the context of visual adversarial threats. The code is available at \url{https://github.com/btzyd/qava}.
\end{abstract}

\section{Introduction}
\label{sec:introduction}

\begin{figure}[ht]
\begin{center}
\centerline{\includegraphics[width=\linewidth]{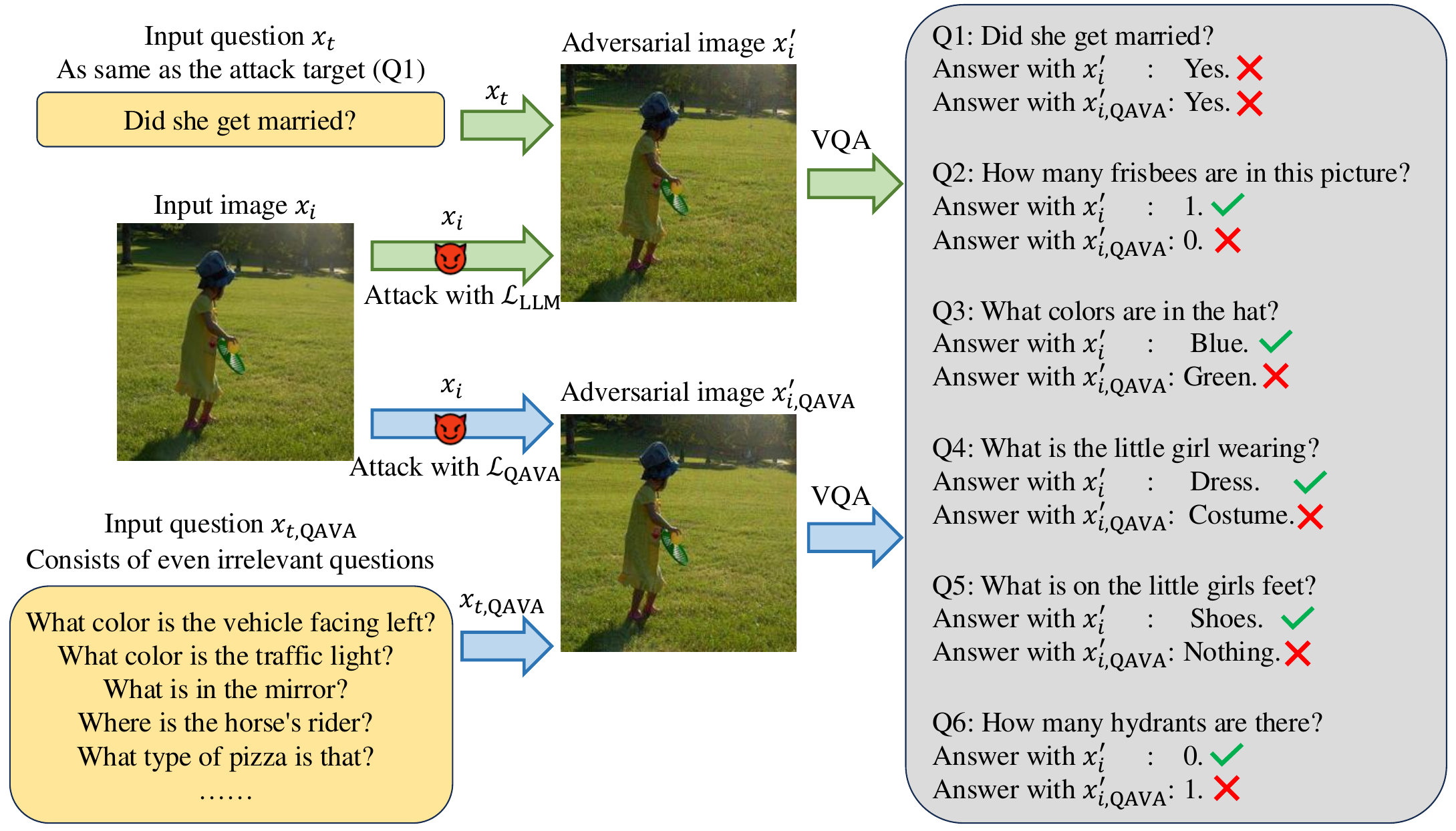}}
\caption{Traditional adversarial attacks involve inputting an image $x_i$ and a specified target question $x_{t,target}$ into LVLMs, with adversarial images generated through gradient-based methods. This approach typically results in incorrect answers for $x_i$ and $x_{t,target}$ (\ie, Q1). However, for other questions $x_{t,other}\in\{x_{t,other}\in\mathcal{T}|x_{t,other}\neq x_{t,target}\}$ within the question set $\mathcal{T}$ that are not the same as the $x_{t,other}$, it remains possible for LVLMs to provide correct answers (\ie, Q2-Q6). Our QAVA samples a set of questions $x_\text{t,QAVA}$ and performs attacks on these questions, even if they are unrelated to the original image $x_i$. QAVA generates adversarial images that are likely to produce incorrect responses when faced with unknown target questions.}
\label{fig:overall}
\end{center}

\end{figure}

With the expansion in model parameters and training datasets, large vision-language models (LVLMs) have gained significant popularity, demonstrating exceptional performance across various tasks, including image classification, image captioning, semantic segmentation, and visual question answering (VQA) \cite{liu2023llava, alayrac2022flamingo, wang2023visionllm}. However, training LVLMs from the ground up is resource-intensive. As a result, the prevailing approach involves fine-tuning pre-trained visual encoders and large language models (LLMs) while training a vision-language alignment module. This process adapts visual tokens to the input space of the LLMs after they pass through the alignment module, enabling the LLMs to effectively process visual tokens. A well-known and efficient visual-language alignment module is Q-former \cite{li2023blip}, which is employed by many popular LVLMs to bridge the visual encoder and the LLM \cite{dai2305instructblip, zhu2023minigpt}.

Despite their robust capabilities, LVLMs remain vulnerable to adversarial attacks. Attack-Bard \cite{dong2023robust} employs surrogate models to manipulate images, causing LVLMs to err on the image captioning task using the fixed prompt, ``Describe this image''. Similarly, VLAttack \cite{yin2023vlattack} targets both visual and textual modalities to disrupt the output of LVLMs for a specific question and image. However, these methods focus exclusively on attacks against \emph{one image and one question at a time}. Consequently, the adversarial examples generated by these attacks may not be effective when confronted with different questions.

To develop more potent attacks, our objective is to manipulate images so that they yield incorrect answers to an \emph{unknown set of target questions}, noted as \textbf{query-agnostic visual attack (QAVA)}. These query-agnostic adversarial examples have the potential to cause significant disruption to the model. For instance, during the inference phase, LVLMs may consistently provide incorrect answers to any question regarding the manipulated image. Moreover, employing these adversarial examples during the training or supervised fine-tuning phase could be even more detrimental, particularly when used for data poisoning. We enhance the attack by scrutinizing both the attack's location and the selection of questions used.

In terms of attack positioning, traditional attack methods typically employ the end-to-end loss function of the entire LVLM as the attack objective function. However, for adversarial attacks, it is crucial to identify and target the most vulnerable component of the LVLM. Given the extensive number of parameters in large language models, we posit that end-to-end attacks on LVLMs may not be as effective as targeting the inputs to the LLMs. Consequently, we focus on \textbf{attacking the output of the visual-language alignment module}. The visual-language alignment module's output encompasses multimodal interactions, and we can disrupt these critical multimodal interactions by targeting the alignment module, potentially leading to more effective attack outcomes.

Regarding the questions employed in attacks, we observed that when targeting the visual-language alignment module, effective attack performance can be achieved even with the use of \emph{randomized questions that are unrelated to the image}. Furthermore, the attack is enhanced when a larger number of random, irrelevant questions are utilized, which verifies our QAVA's flexibility and effectiveness.

In conclusion, QAVA diverges from traditional attacks in two key aspects: (1) We focus on attacking the output of the visual-language alignment module within LVLMs, which is verified to be more vulnerable to query-agnostic attacks. (2) We utilize multiple randomized, image-independent questions in our attacks, ensuring that the adversarial examples are maximally incorrect when confronted with unknown potential inputs.

Our main contributions are summarized as follows: (1) We introduce a query-agnostic attack method, QAVA, which enhances the practicality of adversarial attacks on images within LVLMs. (2) We identify the vulnerability of visual-language alignment modules in LVLMs to adversarial attacks and leverage this vulnerability to execute query-agnostic attacks. (3) Extensive experiments demonstrate the efficacy of our QAVA approach in both white-box and black-box attack scenarios. Additionally, our QAVA method exhibits inter-task transferability, such as transferring from the VQA task to the image captioning task. This serves as an important alert regarding the security of LVLMs.

\section{Related Work}
\label{sec:ro}

\noindent
\textbf{Large vision-language models.}
\label{sec:LVLM}
LVLMs are typically composed of a pre-trained LLM, a visual encoder, and a projector that aligns visual and textual modalities. Recent popular LVLMs include InstructBLIP \cite{dai2305instructblip} and MiniGPT-4 \cite{zhu2023minigpt}. Both models utilize EVA-CLIP \cite{sun2023eva} as the visual encoder and employ the Q-Former \cite{li2023blip} for aligning textual and visual modalities. For the LLM component, models such as Vicuna \cite{chiangvicuna} and FlanT5 \cite{chung2022scaling} are viable options. These LVLMs have demonstrated outstanding performance across various multimodal tasks, including image classification and VQA \cite{antol2015vqa}, among others.

\noindent
\textbf{Adversarial Attacks.}
By introducing small perturbations to the inputs of neural networks, adversarial attacks \cite{szegedy2014intriguing, nguyen2015deep} can cause models to produce incorrect outputs. These attacks can be categorized into white-box and black-box (or gray-box) attacks \cite{papernot2016limitations}. In white-box attacks, the adversary has full access to the model's parameters. Conversely, in black-box or gray-box attacks, the adversary has limited information, such as the ability to make a certain number of queries to the model or knowledge of some of the model's parameters. Furthermore, adversarial attacks can be classified as either targeted or untargeted. Untargeted attacks aim to generate incorrect outputs, while targeted attacks strive to manipulate the output to meet the adversary's specific expectations. 
Initial research on adversarial attacks concentrated mainly on the visual modality, given its high-dimensional and continuous input space \cite{moosavi2016deepfool, goodfellow2015explaining, carlini2017towards}. More recent studies have extended the attacks to discrete textual modalities \cite{alzantot2018generating, jia2017adversarial, wallace2019universal}. Additionally, some research has focused on targeting the fusion of visual and textual modalities \cite{zhang2022towards, lu2023set}.

\noindent
\textbf{LVLMs and Adversarial Attacks.}
With the increasing popularity of LVLMs, numerous recent studies have focused on adversarial attacks against these models. Recent research has demonstrated the feasibility of generating adversarial examples to jailbreak LVLMs \cite{shayegani2023survey}. This includes attacking images using gradient-based approaches \cite{carlini2023aligned}, targeting texts through prompt engineering \cite{liu2023jailbreaking}, and embedding malicious instructions into images as text, with the aim of having the model execute these commands via optical character recognition (OCR) \cite{shayegani2023jailbreak}. While these studies primarily address the security concerns surrounding LVLMs, our research is specifically focused on the safety and integrity of images within these models. Some studies \cite{luo2024an} have also concentrated on query-agnostic adversarial attacks, in which LVLMs are prompted to respond with answers such as ``none'' or ``don't know'' to various inquiries. In contrast, our study specifically examines scenarios in which LVLMs are induced to provide incorrect answers.

\begin{figure}[t]
    \begin{center}
    \centerline{\includegraphics[width=\linewidth]{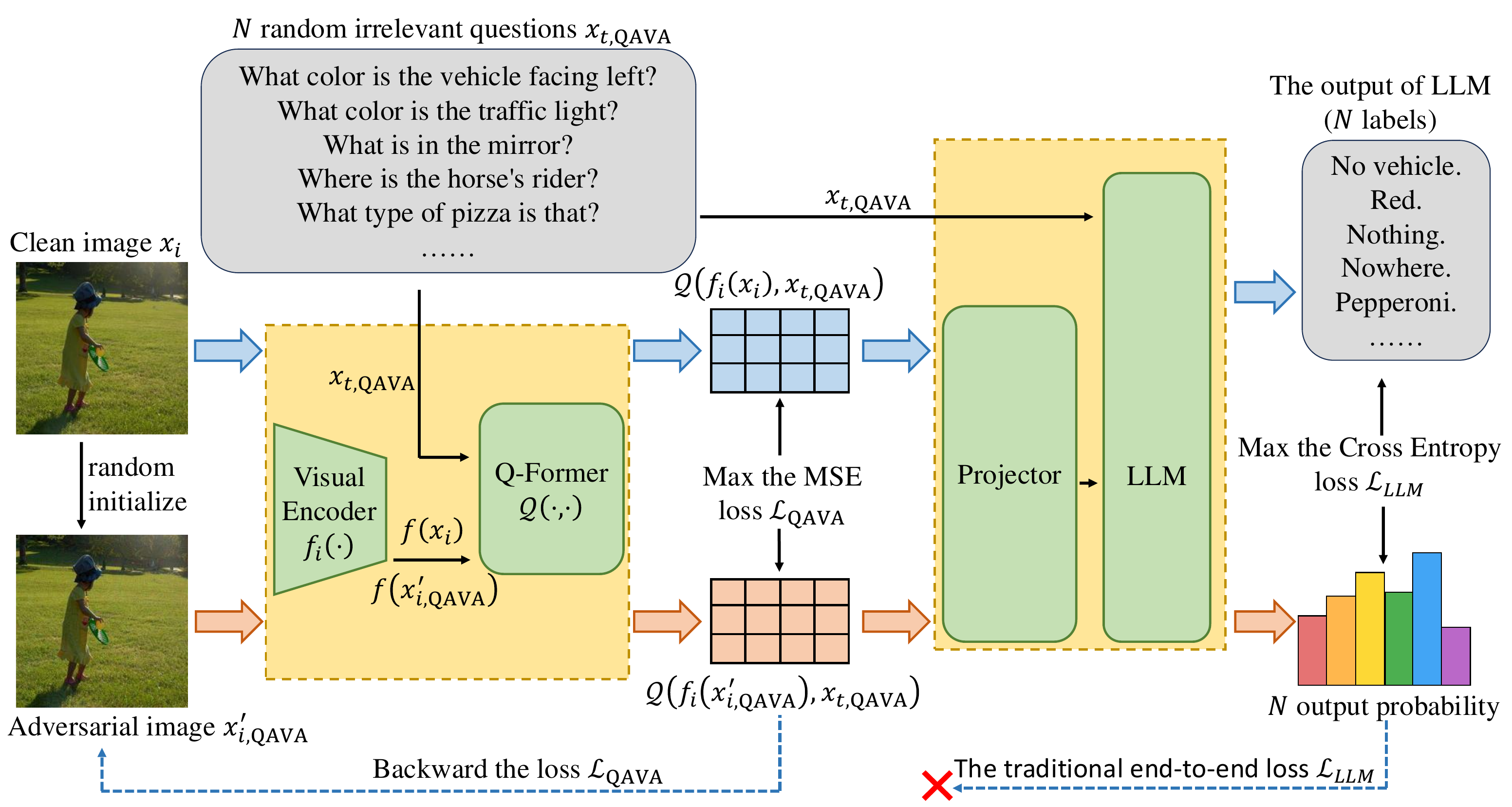}}
    \caption{The framework of QAVA is structured as follows: Initially, we generate $N$ randomly sampled questions, denoted as $x_\text{t,QAVA}$, which are not pertinent to the input image $x_i$. Subsequently, we introduce random perturbations to $x_i$ to create the initial variant, $x'_\text{i,QAVA}$. Both $x_i$ and $x_\text{t,QAVA}$ are then input into the LVLM, and the LVLM's response serves as a label. Despite the fact that the question $x_\text{t,QAVA}$ is unrelated to the image $x_i$, the LVLM still provides a response. Following this, we input $x'_\text{i,QAVA}$ and $x_\text{t,QAVA}$ into the LVLM to calculate the MSE loss based on the Q-former output features. Adversarial attacks are executed using techniques such as PGD or C\&W by employing the loss functions, denoted as $\mathcal{L}_\text{QAVA}$. The traditional end-to-end attack loss function, $\mathcal{L}_\text{LLM}$, is also shown.}
    \label{fig:overall2}
    \end{center}
\end{figure}

\section{Method}
\label{sec:method}

\subsection{Preliminary}
\label{sec:preliminary}

We provide a concise overview of the adversarial attack pipeline. This study specifically investigates gradient-based white-box adversarial attacks. Let the LVLM be represented as $y = f(f_i(x_i), x_t)$, where $x_i$ and $x_t$ denote the input image and text, respectively, with $f_i(\cdot)$ serving as a visual encoder, and $y$ as the textual output generated by the LVLM. Given the input image $x_i$ and text $x_t$, our objective is to identify an adversarial image $x'_{i}$ such that $y' = f(f_i(x'_i), x_t)$ is semantically distant from $y$. This is subject to the condition that the difference between $x_i$ and $x'_i$ remains within the constraints $\epsilon$, \ie, ${\left|x'_{i} - x_i \right|}_p \leq \epsilon$.

FGSM \cite{goodfellow2015explaining} generates the adversarial example $x'_i$ by updating the original input $x_i$ using a single gradient computation. In contrast, PGD \cite{madry2018towards} executes multiple iterative gradient updates, projecting $x'_i$ after each update to ensure adherence to the perturbation constraints defined by $\epsilon_\infty$. The C\&W attack \cite{carlini2017towards} employs \cref{eq:cw} as its optimization objective, which is iterated multiple times. The first component of \cref{eq:cw} seeks to modify the model output to diverge significantly from the original output by employing a specific loss function $\mathcal{L}$. Meanwhile, the second component ensures that the adversarial example $x'$ remains sufficiently close to the original input $x$. The constant $c$ serves as a hyperparameter, balancing the divergence in model outputs against the $l_2$ distance between $x'$ and $x$.
\begin{equation}
    \label{eq:cw}
    \mathcal{L}_{CW}(x, x', \dots) = \mathcal{L}(x, x', \dots) - c\times||x-x'||_2^2
\end{equation}

The objective of QAVA is to manipulate images such that they yield incorrect responses to unknown target questions. Consequently, our task involves employing a specific method to adversarially attack a given image. We explore the selection of surrogate questions for these attacks in \cref{sec:rsq} and detail the associated loss functions in \cref{sec:loss}.

\subsection{Strategies for Sampling Questions}
\label{sec:rsq}
We outline the four question sampling strategies utilized in our QAVA attack as follows.

\noindent
\textbf{White-box targeting questions (WTQ).}  WTQ employs a predefined set of target questions that are used to evaluate the adversarial example generated by attack. In contrast, our QAVA method does not have access to any information regarding the target questions at the time of the attack.

\noindent
\textbf{Visual question generation (VQG).} VQG \cite{mostafazadeh2016generating} is capable of generating questions for input images, including questions with specific anticipated answers (\eg, ``yes'' or ``green''). In this strategy, we input the images into LVLMs and generate $N$ questions using VQG prompts (\eg, ``Taking the image into account, generate $N$ questions.'').

\noindent
\textbf{Random sample questions ($\text{RSQ}_\text{N}$).} $\text{RSQ}_\text{N}$ randomly samples $N$ questions from the validation set of VQA v2, which comprises 214,354 questions.

\noindent
\textbf{Random Sample Questions by Types ($\text{RSQ}^\text{t}$).} $\text{RSQ}^\text{t}$ ensures a balanced representation of each question type in the final set of sampled questions. This approach involves categorizing questions by type, such as ``What is on the'', ``What animal is'', ``What color is'', among others.

\subsection{Design of the Loss Function $\mathcal{L}_\text{QAVA}$}
\label{sec:loss}
For the LVLM $f$, the forward process $f(f_i(x_i), x_t)$ with a label generates a native loss function, $\mathcal{L}_\text{LLM}$, which is typically employed to optimize the adversarial image. Our aim is to identify more effective loss functions to enhance attack performance.

Revisiting the forward process of LVLMs, the Q-former, introduced by BLIP-2 and utilized in LVLMs such as InstructBLIP and MiniGPT-4, is instrumental in aligning visual and textual modalities within the feature space. For instance, in InstructBLIP, the image $x_i$ is encoded into a vector of shape $[257, 1408]$, denoted as $f_i(x_i)$, by the image encoder $f_i$. The Q-former $\mathcal{Q}(\cdot, \cdot)$ then extracts a feature of shape $[32, 768]$, represented as $q = \mathcal{Q}(f_i(x_i), x_t)$, guided by the text $x_t$. This feature $q$ is subsequently upscaled to $[32, 4096]$ and input into the LLM along with the text $x_t$, as illustrated in \cref{fig:overall2}.

The feature vectors output by the Q-former can be utilized as supervised signals to optimize the adversarial image. Specifically, for a given question $x_t$ and clean image $x_i$, we first input them into the Q-former to obtain $q = \mathcal{Q}(f_i(x_i), x_t)$. Next, we input the question $x_t$ and the perturbed image $x'_i$ into the Q-former to derive $q' = \mathcal{Q}(f_i(x'_i), x_t)$. We then optimize the MSE loss functions, \ie
\begin{align}
    &\text{max}\quad\mathcal{L}_\text{QAVA}(q,q')=\frac{1}{MN}\sum_{i=1}^{M}\sum_{j=1}^{N}\left(q_{i,j}-q'_{i,j}\right)^2 \nonumber\\
    &\text{subject to}\quad|x_i-x'_i|_\infty\leq\epsilon_\infty
\end{align}

The loss function $\mathcal{L}_\text{QAVA}$ is calculated for all white-box questions (\eg, WTQ) or surrogate questions (\eg, RSQ). After summing these individual computations, we obtain the overall loss function $\mathcal{L}_\text{QAVA}$, as detailed in \cref{alg:example} in \cref{sec:algorithm}.

\section{Experiments}
\label{sec:experiments}

\subsection{Experiment Settings}
\label{sec:experiment_settings}

\noindent
\textbf{Datasets.} We utilize the validation set of VQA v2 as the foundational dataset, comprising a total of 40,504 images and 214,354 questions. The distribution of questions across images in VQA v2 is uneven; for instance, over 18,000 images have only three questions, whereas merely 50 images have 50 or more questions. We define the dataset VQA v2 $m$+$n$ as a subset of VQA v2, including $m$ images, each associated with $n$ questions, resulting in a total of $m \times n$ questions. To construct the dataset VQA v2 $m$+$n$, we randomly sample $m$ images from those with at least $n$ corresponding questions, followed by randomly selecting $n$ questions for each image. Our experiments were performed on the VQA v2 32+50 data subset. We adhere to the official evaluation procedure designated for the VQA v2 dataset. It is crucial to acknowledge that the ground truth answers for each question in VQA v2 are not singular; each validation question possesses ten ground truth answers, which are utilized to compute the VQA scores.

\noindent
\textbf{Models.} In our experiments, we employ BLIP-2 \cite{li2023blip} and InstructBLIP \cite{dai2305instructblip}. Both LVLMs utilize CLIP as the visual encoder and incorporate a set of learnable queries along with a Q-former trained on a frozen visual encoder and LLM. Additionally, we assess the transferability of our QAVA approach on LLaVA \cite{liu2023llava} and MiniGPT-4 \cite{zhu2023minigpt}.

\noindent
\textbf{Adversarial Attacks.} We employ two standard adversarial attack methods: $\text{PGD-}{l_\infty}$ and $\text{CW-}{l_2}$. For the PGD attack, unless otherwise specified, we typically set the number of attack steps $n$ to 20, the attack step size $\alpha$ to 2 (\ie, $2/255$), and the maximum perturbation magnitude $\epsilon_\infty$ to 8. For the CW attack, we generally configure the number of attack steps $n$ to 50, the attack step size $\alpha$ to 0.01 (\ie, approximately $2.55/255$), and set the confidence level to 0. The choice of the constant $c$ in the CW attack is contingent upon the loss function employed. Specifically, since $\mathcal{L}_\text{LLM}$ is approximately 20 times larger than $\mathcal{L}_\text{QAVA}$, we set the constant $c=0.1$ when using $\mathcal{L}_\text{LLM}$ as the loss function. Conversely, when utilizing $\mathcal{L}_\text{QAVA}$, we set the constant $c=0.005$.

\subsection{Main Results of QAVA}
\label{sec:results_rsq}

\noindent
\textbf{$\mathcal{L}_\text{LLM}$($\text{RSQ}_\text{N}$) can effectively attack images}. We assess the baseline VQA performance of InstructBLIP with $\text{FlanT5}_\text{XL}$ and InstructBLIP with Vicuna-7B on the VQA v2 32+50 dataset. Subsequently, we apply PGD and CW attacks using WTQ to evaluate the maximum potential of attack performance. Following this, we randomly select $N$ questions from the entire set of VQA v2 questions (214,354 questions) to form $\text{RSQ}_\text{N}$ and use these for the attack. The results, presented in \cref{tab:rsq_loss_llm}, demonstrate that both PGD and CW can effectively reduce VQA scores when white-box questions are used. However, when using randomly sampled irrelevant questions, PGD and CW do not reach the same level of performance as with WTQ, displaying a difference of approximately 10 to 15 points.

\begin{table}[ht]
    \centering
    \resizebox{\linewidth}{!}{
    \begin{tabular}{c|c|c|c|c|c}
        \toprule
        Attack & Question & \multicolumn{4}{c}{InstructBLIP(Vicuna-7B)}\\
        method & strategy & Overall & Other & Number & Yes/No \\
        \midrule
        \rowcolor{gray!30}\ding{56} & \ding{56} & 78.00 & 66.98 & 69.68 & 94.29 \\ 
        \midrule
        \multirow{7.44}{*}{PGD} & $\text{WTQ}_\text{50}$ & 42.41 & 17.56  & 50.59 & 71.13\\
        \cmidrule{2-6}
        & $\text{RSQ}_\text{1}$ & 62.81{\scriptsize ($\pm$1.34)} & 45.20  & 60.13 & 85.68 \\
        & $\text{RSQ}_\text{5}$ & 58.46{\scriptsize ($\pm$0.66)} & 39.72  & 56.37 & 82.58 \\
        & $\text{RSQ}_\text{10}$ & 57.61{\scriptsize ($\pm$1.50)} & 37.71  & 55.24 & 83.28 \\
        & $\text{RSQ}_\text{15}$ & 55.08{\scriptsize ($\pm$1.85)} & 34.39  & 53.22 & 81.57 \\
        & $\text{RSQ}_\text{20}$ & 55.42{\scriptsize ($\pm$0.72)} & 34.58  & 55.42 & 81.55 \\
        & $\text{RSQ}_\text{25}$ & 54.53{\scriptsize ($\pm$1.40)} & 33.65  & 54.26 & 80.79\\
        \midrule
        \multirow{7.44}{*}{CW} & $\text{WTQ}_\text{50}$ & 40.66 & 16.93  & 43.05 & 69.68\\
        \cmidrule{2-6}
        & $\text{RSQ}_\text{1}$ & 61.30{\scriptsize ($\pm$1.73)} & 44.65  & 58.45 & 83.03 \\
        & $\text{RSQ}_\text{5}$ & 58.58{\scriptsize ($\pm$0.63)} & 40.74  & 56.20 & 81.65 \\
        & $\text{RSQ}_\text{10}$ & 57.65{\scriptsize ($\pm$1.33)} & 39.24  & 53.22 & 82.05 \\
        & $\text{RSQ}_\text{15}$ & 57.05{\scriptsize ($\pm$1.99)} & 38.01  & 54.28 & 81.74 \\
        & $\text{RSQ}_\text{20}$ & 57.13{\scriptsize ($\pm$0.95)} & 37.42  & 55.79 & 82.25 \\
        & $\text{RSQ}_\text{25}$ & 55.68{\scriptsize ($\pm$1.42)} & 35.64  & 56.03 & 80.69\\
        \bottomrule
    \end{tabular}
    }
    \caption{$\mathcal{L}_\text{LLM}$($\text{RSQ}_\text{N}$) can effectively attack images, but its performance still lags behind that of WTQ.}
    \label{tab:rsq_loss_llm}
\end{table}

\noindent
\textbf{$\mathcal{L}_\text{QAVA}$($\text{RSQ}_\text{N}$) can further improve attack performance than $\mathcal{L}_\text{LLM}$($\text{RSQ}_\text{N}$)}. $\mathcal{L}_\text{LLM}$($\text{RSQ}_\text{N}$) approach typically employs the end-to-end loss function $\mathcal{L}_\text{LLM}$ of the LLMs. In contrast, we propose targeting the visual-language alignment module using $\mathcal{L}_\text{QAVA}$, as defined in \cref{sec:loss}. We compared the effectiveness of $\mathcal{L}_\text{LLM}$ and $\mathcal{L}_\text{QAVA}$, with results presented in \cref{tab:rsq_loss_QAVA}. For WTQ, there is no significant difference in performance between $\mathcal{L}_\text{QAVA}$ and $\mathcal{L}_\text{LLM}$. However, for RSQ, \emph{\textbf{using $\mathcal{L}_\text{QAVA}$ results in a significantly better attack performance than $\mathcal{L}_\text{LLM}$}}. This highlights the effectiveness of our approach $\mathcal{L}_\text{QAVA}$($\text{RSQ}_\text{N}$), indicating its potential to enhance adversarial attack capabilities.

\begin{table}[ht]
    \centering
    \resizebox{\linewidth}{!}{
    \begin{tabular}{c|c|c|c|c|c|c}
        \toprule
        Attack & Loss & Question & \multicolumn{4}{c}{InstructBLIP(Vicuna-7B)}\\
        method & $\mathcal{L}$ & strategy & Overall & Other & Number & Yes/No \\
        \midrule
        \rowcolor{gray!30}\ding{56} & \ding{56} & \ding{56} & 78.00 & 66.98 & 69.68 & 94.29 \\ 
        \midrule
        \multirow{7.34}{*}{PGD} & \multirow{2}{*}{$\mathcal{L}_\text{LLM}$} & $\text{WTQ}_\text{50}$ & 42.41 & 17.56 & 50.59 & 71.13 \\
        & & $\text{RSQ}_\text{25}$ & 54.53{\scriptsize ($\pm$1.40)} & 33.65 & 54.26 & 80.79 \\
        \cmidrule{2-7}
        & \multirow{5}{*}{$\mathcal{L}_\text{QAVA}$} & $\text{WTQ}_\text{50}$ & 44.41 & 21.12  & 43.96 & 73.75 \\
        & & $\text{RSQ}_\text{1}$ & 45.70{\scriptsize ($\pm$1.06)} & 21.70  & 46.29 & 75.60 \\
        & & $\text{RSQ}_\text{5}$ & 43.59{\scriptsize ($\pm$1.16)} & 20.76  & 42.00 & 72.69 \\
        & & $\text{RSQ}_\text{10}$ & 44.85{\scriptsize ($\pm$1.07)} & 22.11  & 43.25 & 73.83 \\
        & & $\text{RSQ}_\text{25}$ & 44.07{\scriptsize ($\pm$0.83)} & 20.15  & 46.60 & 73.32 \\
        \midrule
        \multirow{7.34}{*}{CW} & \multirow{2}{*}{$\mathcal{L}_\text{LLM}$} & $\text{WTQ}_\text{50}$ & 40.66 & 16.93 & 43.05 & 69.68 \\
        & & $\text{RSQ}_\text{25}$ & 55.68{\scriptsize ($\pm$1.42)} & 35.64 & 56.03 & 80.69 \\
        \cmidrule{2-7}
        & \multirow{5}{*}{$\mathcal{L}_\text{QAVA}$} & $\text{WTQ}_\text{50}$ & 41.82 & 17.63  & 39.68 & 72.79 \\
        & & $\text{RSQ}_\text{1}$ & 43.16{\scriptsize ($\pm$1.03)} & 19.39  & 42.24 & 73.25 \\
        & & $\text{RSQ}_\text{5}$ & 42.18{\scriptsize ($\pm$1.15)} & 18.37  & 41.48 & 72.24 \\
        & & $\text{RSQ}_\text{10}$ & 41.85{\scriptsize ($\pm$0.83)} & 17.38  & 41.03 & 72.77 \\
        & & $\text{RSQ}_\text{25}$ & 40.98{\scriptsize ($\pm$1.71)} & 16.45  & 40.48 & 71.88 \\
        \bottomrule
    \end{tabular}
    }
    \caption{$\mathcal{L}_\text{QAVA}$ is better than $\mathcal{L}_\text{LLM}$ in use of $\text{RSQ}_\text{N}$.}
    \label{tab:rsq_loss_QAVA}
\end{table}

\noindent
\textbf{Generalizability of QAVA to other LVLMs}. Previously, our experiments focused solely on the InstructBLIP Vicuna-7B. To assess the broader applicability of QAVA, we extended our evaluation to include additional LVLMs, with results summarized in \cref{tab:more_lvlms_qava}. The findings demonstrate that QAVA consistently delivers effective attack performance across a wider spectrum of LVLMs, showcasing its robustness and adaptability in diverse LVLMs.

\begin{table}[ht]
    \centering
    \resizebox{\linewidth}{!}{
    \begin{tabular}{c|c|c|c}
        \toprule
         Model & Clean & $\mathcal{L}_\text{LLM}(\text{RSQ}_{25})$ & $\mathcal{L}_\text{QAVA}(\text{RSQ}_{10})$ \\
         \midrule
        BLIP-2 opt-2.7B & \cellcolor{gray!30}45.64 & 34.56 & 19.15 \\
        BLIP-2 $\text{FlanT5}_\text{XL}$ & \cellcolor{gray!30}62.05 & 29.39 & 32.28 \\
        BLIP-2 opt-6.7B & \cellcolor{gray!30}48.26 & 15.25 & 19.49 \\
        BLIP-2 $\text{FlanT5}_\text{XXL}$ & \cellcolor{gray!30}62.10 & 29.11 & 26.93 \\
        InstructBLIP $\text{FlanT5}_\text{XL}$ & \cellcolor{gray!30}74.54 & 49.78 & 34.31 \\
        InstructBLIP Vicuna-7B & \cellcolor{gray!30}78.00 & 54.53 & 44.85 \\
        InstructBLIP $\text{FlanT5}_\text{XXL}$ & \cellcolor{gray!30}73.02 & 48.57 & 34.32 \\
        InstructBLIP Vicuna-13B & \cellcolor{gray!30}67.87 & 53.19  & 42.90 \\
        \bottomrule
    \end{tabular}
    }
    \caption{Results of QAVA on various LVLMs.}
    \label{tab:more_lvlms_qava}
\end{table}

\noindent
\textbf{Generalizability of QAVA to other datasets}. To further assess the applicability of QAVA, we conducted evaluations using the VizWiz test-dev dataset \cite{gurari2018vizwiz, gurari2019vizwiz}, which consists of 8,000 image-question pairs, each image paired with a single question. We explored two attack scenarios: $\mathcal{L}_\text{LLM}(\text{WTQ}_{1})$, which represents a traditional end-to-end adversarial attack targeting the specific question, and $\mathcal{L}_\text{QAVA}(\text{RSQ}_{10})$, which illustrates the QAVA attack using 10 randomly selected questions from VQA v2 without prior knowledge of the target questions. As shown in \cref{tab:more_datasets_qava}, QAVA reliably performs effective adversarial attacks using these randomized questions against unknown target questions on VizWiz. These results underscore the versatility and robustness of QAVA across different datasets and question distributions.

\begin{table}[ht]
    \centering
    \resizebox{\linewidth}{!}{
    \begin{tabular}{c|c|c|c|c|c}
        \toprule
         Attack & \multicolumn{5}{c}{VizWiz test-dev evaluation} \\
         method & overall & yes/no & number & other & unanswerable \\
         \midrule
         \cellcolor{gray!30}Clean & \cellcolor{gray!30}33.08 & \cellcolor{gray!30}81.74 & \cellcolor{gray!30}28.10 & \cellcolor{gray!30}39.14 & \cellcolor{gray!30}10.99 \\
         $\mathcal{L}_\text{LLM}(\text{WTQ}_{1})$ & 10.48 & 63.71 & 8.25 & 8.44 & 6.93 \\
         $\mathcal{L}_\text{QAVA}(\text{RSQ}_{10})$ & 10.69 & 59.04 & 7.78 & 9.54 & 5.85 \\
        \bottomrule
    \end{tabular}
    }
    \caption{Results of QAVA on VizWiz test-dev dataset.}
    \label{tab:more_datasets_qava}
\end{table}

\subsection{Ablation Study of Question Sampling}
\label{sec:ablation_vqg}

\textbf{No need for image-related surrogate questions}. VQG for images requires LVLMs to process the image and generate numerous tokens, which can be resource-intensive and time-consuming. While this approach ensures that the questions used for the attack are closely related to the image, the results presented in \cref{tab:VQG} indicate that this does not lead to a significant improvement in attack performance. Consequently, in practical applications, it is unnecessary to employ VQG to create the set of surrogate questions, as the benefits in terms of attack efficacy do not justify the additional computational cost.

\begin{table}[ht]
    \centering
    \resizebox{\linewidth}{!}{
    \begin{tabular}{c|c|c|c|c|c}
        \toprule
         Attack & Question & \multicolumn{4}{c}{VQA v2 scores} \\
         method & strategy & Overall & Other & Number & Yes/No  \\
        \midrule
         \multirow{5.4}{*}{PGD} & $\text{WTQ}_\text{50}$ & 44.41 & 21.12 & 43.96 & 73.75 \\
         \cmidrule{2-6}
         & $\text{RSQ}_\text{10}$ & 44.85{\scriptsize ($\pm$1.07)} & 22.11  & 43.25 & 73.83 \\
         \cmidrule{2-6}
         & $\text{RSQ}^\text{t}_\text{10}$ & 43.85{\scriptsize ($\pm$1.29)} & 20.32  & 44.54 & 73.13 \\
         \cmidrule{2-6}
         & $\text{VQG}_\text{10}$ & 43.26{\scriptsize ($\pm$0.67)} & 20.83  & 39.44 & 72.53 \\
        \midrule
         \multirow{5.4}{*}{CW} & $\text{WTQ}_\text{50}$ & 41.82 & 17.63 & 39.68 & 72.79 \\
         \cmidrule{2-6}
         & $\text{RSQ}_\text{10}$ & 41.85{\scriptsize ($\pm$0.83)} & 17.38  & 41.03 & 72.77 \\
         \cmidrule{2-6}
         & $\text{RSQ}^\text{t}_\text{10}$ & 42.05{\scriptsize ($\pm$1.22)} & 17.79  & 40.22 & 73.01\\
         \cmidrule{2-6}
         & $\text{VQG}_\text{10}$ & 39.38{\scriptsize ($\pm$1.07)} & 16.00  & 38.10 & 69.08 \\
        \bottomrule
    \end{tabular}
    }
     \caption{The ablation study of the question sampling strategy as outlined in \cref{sec:rsq}. In all experiments, the loss function $\mathcal{L}_\text{QAVA}$ is utilized on InstructBLIP Vicuna-7B. $\text{RSQ}_\text{10}$ involves the random sampling of 10 questions from the entire list of available questions. In contrast, $\text{RSQ}^\text{t}_\text{10}$ first randomly selects 10 question types from the 67 types identified in VQA v2 and subsequently samples one question from each selected type, ensuring that no more than one question per type is sampled. The approach $\text{VQG}_\text{10}$ employs MiniGPT-4 to generate 10 questions specifically for the images.}
     \label{tab:VQG}
\end{table}

\noindent
\textbf{Randomly sampling questions is simple and efficient.} The findings in \cref{tab:VQG} show that $\text{RSQ}^\text{t}$ does not significantly enhance attack performance. Consequently, a straightforward approach of randomly sampling questions $\text{RSQ}_\text{N}$ is sufficient and effective. This method not only simplifies the process but also reduces computational overhead, all while maintaining robust attack performance.

\subsection{Abalation Study of the Imperceptibility of Images Generated by QAVA}
\label{sec:abalation_imperceptibility}

\begin{figure}[ht]
    \centering
    \begin{subfigure}{0.49\linewidth}
        \includegraphics[width=\linewidth]{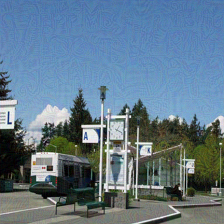}
        \caption{The QAVA image.}
        \label{fig:imperceptibility_small_qava}
    \end{subfigure}
    \begin{subfigure}{0.49\linewidth}
        \includegraphics[width=\linewidth]{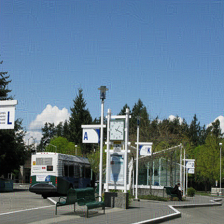}
        \caption{The QAVA+SSAH image.}
        \label{fig:imperceptibility_small_qava_ssah}
    \end{subfigure}
    \caption{Visualization of image imperceptibility.}
    \label{fig:figure_imperceptibility_small}
\end{figure}

Although we employed a smaller attack strength (\eg, $\epsilon_\infty=8$ in PGD), adversarial examples may still be detectable upon careful examination, as shown in \cref{fig:imperceptibility_small_qava}. Previous studies have developed techniques to enhance image imperceptibility in classical attack methods, such as SSAH \cite{luo2022frequency}. By integrating QAVA with SSAH, we can produce adversarial images that are both imperceptible and query-agnostic, as demonstrated in \cref{fig:imperceptibility_small_qava_ssah}. The VQA scores for QAVA and QAVA+SSAH are demonstrated in \cref{tab:qava_ssah}. There exists a trade-off between the imperceptibility of adversarial examples and the effectiveness of adversarial attacks. Nevertheless, QAVA+SSAH successfully generates adversarial examples that are both imperceptible and exhibit a significant attack impact. Additional adversarial images are presented in \cref{fig:figure_imperceptibility}. Furthermore, we investigate the effect of varying attack strengths on the efficacy of our approach, as detailed in \cref{sec:ablation_attack_methods}.

\begin{table}[ht]
    \centering
    \resizebox{\linewidth}{!}{
    \begin{tabular}{c|c|c|c|c}
        \toprule
        \multirow{2}{*}{Attack method} & \multicolumn{4}{c}{VQA v2 scores} \\
         & Overall & Other & Number & Yes/No \\
        \midrule
        QAVA & 44.85 & 22.11 & 43.25 & 73.83 \\
        QAVA+SSAH & 50.67 & 28.64 & 50.37 & 78.37 \\
        \bottomrule
    \end{tabular}
    }
    \caption{The combination of QAVA and SSAH generates imperceptible adversarial examples.}
    \label{tab:qava_ssah}
\end{table}

\subsection{Ablation Study of QAVA Attack Strength}
\label{sec:ablation_attack_methods}

Our experiments concentrate on two main attack methods: PGD and C\&W, as described in \cref{sec:experiment_settings}. To evaluate the balance between attack efficacy and image imperceptibility, we investigated different levels of attack strength. As demonstrated in \cref{tab:qava_strength}, our QAVA approach consistently generates effective adversarial examples across various attack strengths. Even at lower attack intensities, there is a notable reduction in VQA scores, while the noise introduced remains nearly imperceptible.

\begin{table}[ht]
    \centering
    \resizebox{\linewidth}{!}{
    \begin{tabular}{c|c|c|c|c|c}
        \toprule
        Attack & $\epsilon_\infty$ of PGD & \multicolumn{4}{c}{VQA v2 scores} \\
        method & $c$ of C\&W &  Overall & Other & Number & Yes/No \\
        \midrule
        \multirow{3}{*}{PGD} & 4/255 &  48.93 & 27.57 & 46.42 & 76.46 \\
        & 8/255 & 44.85 & 22.11 & 43.25 & 73.83 \\
        & 16/255 & 38.86 & 16.02 & 38.29 & 67.67 \\
        \midrule
        \multirow{3}{*}{C\&W} & 0.05 & 51.72 & 32.11 & 48.13 & 77.37 \\
        & 0.005 & 41.85 & 17.38 & 41.03 & 72.77 \\
        & 0.0005 & 36.44 & 11.41 & 38.45 & 67.22 \\
        \bottomrule
    \end{tabular}
    }
    \caption{The alternative attack strength of $\mathcal{L}_\text{QAVA}$ with $\text{RSQ}_{10}$. The other settings are the same as \cref{sec:experiment_settings}.}
    \label{tab:qava_strength}
\end{table}

\subsection{The Efficiency of QAVA Attacks}
\label{sec:qava_efficiency}

\Cref{tab:rsq_loss_QAVA} illustrates that our QAVA $\mathcal{L}_\text{QAVA}$($\text{RSQ}_{10}$) attains performance levels comparable to traditional end-to-end adversarial attacks method $\mathcal{L}_\text{LLM}$($\text{WTQ}_{50}$) that directly target 50 specific questions. To further underscore the efficiency of our QAVA approach, we evaluate three scenarios: traditional end-to-end attacks targeting 1 question and 50 questions, and QAVA targeting 10 randomly selected questions. In comparison to the traditional end-to-end approach, our QAVA method offers significant improvements in both time and memory efficiency. As indicated in \Cref{tab:qava_efficiency}, QAVA achieves approximately 80\% savings in GPU memory usage because it targets only the output of the vulnerable visual language alignment module, bypassing the resource-intensive LLM. Regarding time efficiency, while the traditional approach is rapid when attacking a single question, it proves ineffective for a series of target questions. Conversely, attacking all 50 questions using the traditional method is effective but requires prior knowledge of all target questions and is time-intensive. In contrast, our QAVA approach, which employs 10 random questions without prior knowledge of specific targets, achieves attack results comparable to those of the traditional method across all target questions, while requiring only 5\% of the time consumed by the traditional end-to-end adversarial attack.

\begin{table}[ht]
    \centering
    \resizebox{\linewidth}{!}{
    \begin{tabular}{c|c|c|c}
        \toprule
         Attack method & Time (s) & GPU Mem (GB) & VQA scores \\
         \midrule
         \cellcolor{gray!30}Clean & \cellcolor{gray!30}\ding{56} & \cellcolor{gray!30}\ding{56} & \cellcolor{gray!30}78.00 \\
        $\mathcal{L}_\text{LLM}$($\text{WTQ}_1$) & \cellcolor{green!40}208 & \cellcolor{red!40}32.62 & \cellcolor{red!40}57.07 \\
        $\mathcal{L}_\text{LLM}$($\text{WTQ}_{50}$) & \cellcolor{red!40}7788 & \cellcolor{red!40}32.62 & \cellcolor{green!40}42.41 \\
        $\mathcal{L}_\text{QAVA}$($\text{RSQ}_{10}$) & \cellcolor{green!40}387 & \cellcolor{green!40}6.30 & \cellcolor{green!40}44.85 \\
        \bottomrule
    \end{tabular}
    }
    \caption{The efficiency of QAVA over traditional end-to-end adversarial attacks. Red cells denote worse performance, while green cells indicate better performance.}
    \label{tab:qava_efficiency}
\end{table}

\subsection{Experiments on QAVA's Transferability}

\begin{table}[ht]
    \centering
    \resizebox{\linewidth}{!}{
    \begin{tabular}{c|c|c}
        \toprule
        \multirow{2}{*}{Surrogate model} & \multicolumn{2}{c}{LLaVA VQA scores} \\
        & $m=0$ & $m=0.9$ \\
         \midrule
        \cellcolor{gray!30}Clean image & \multicolumn{2}{c}{\cellcolor{gray!30}78.32}  \\
        InstructBLIP $\text{FlanT5}_\text{XL}$ & 65.83 & 64.99 \\
        InstructBLIP Vicuna-7B & 68.76 & 64.69 \\ 
        InstructBLIP $\text{FlanT5}_\text{XXL}$ & 66.97 & 66.82 \\
        InstructBLIP Vicuna-13B &67.17 & 63.86 \\
        \bottomrule
    \end{tabular}
    }
    \caption{Results of transfer attacking LLaVA using DI+MI with $\mathcal{L}_\text{QAVA}$($\text{RSQ}_{10}$), where $m$ denotes the momentum. We use the LLaVA-v1.5-7b model with CLIP-ViT-L-336px, which differs significantly from InstructBLIP. Despite this, QAVA still has good transferability.}
    \label{tab:transfer_vqa_to_llava}
\end{table}

\begin{table*}[ht]
    \centering
    \resizebox{\linewidth}{!}{
    \begin{tabular}{c|c|c|c|c|c|c|c|c}
        \toprule
        \multirow{2}{*}{Surrogate model} & \multicolumn{4}{c|}{BLIP-2} & \multicolumn{4}{c}{InstructBLIP} \\
        & opt-2.7B & $\text{FlanT5}_\text{XL}$ & opt-6.7B & $\text{FlanT5}_\text{XXL}$ & $\text{FlanT5}_\text{XL}$ & Vicuna-7B & $\text{FlanT5}_\text{XXL}$ & Vicuna-13B \\
         \midrule
        \cellcolor{gray!30}Clean images & \cellcolor{gray!30}45.64 & \cellcolor{gray!30}62.05 & \cellcolor{gray!30}48.26 & \cellcolor{gray!30}62.10 & \cellcolor{gray!30}74.54 & \cellcolor{gray!30}78.00 & \cellcolor{gray!30}73.02 & \cellcolor{gray!30}67.87 \\
        BLIP-2 opt-2.7B &19.15 &27.63 &19.48 &28.09 &42.37 &44.93 &41.28 &45.84 \\
        BLIP-2 $\text{FlanT5}_\text{XL}$ &27.26 &32.28 &24.68 &35.88 &45.64 &48.1 &46.29 &48.58 \\
        BLIP-2 opt-6.7B &22.47 &28.2 &19.49 &34.22 &45.32 &45.53 &41.69 &44.25 \\
        BLIP-2 $\text{FlanT5}_\text{XXL}$ &20.22 &29.07 &25.07 &26.93 &44.81 &47.35 &44.39 &46.91 \\
        InstructBLIP $\text{FlanT5}_\text{XL}$ &22.99 &28.18 &23.52 &24.29 &34.31 &45.34 &43.35 &47.98 \\
        InstructBLIP Vicuna-7B &20.15 &22.85 &19.35 &27.1 &38.53 &44.85 &38.83 &43.61 \\
        InstructBLIP $\text{FlanT5}_\text{XXL}$ &18.07 &27.43 &16.81 &29.47 &42.67 &46.46 &34.32 &46.84 \\
        InstructBLIP Vicuna-13B &19.85 &22.2 &19.76 &30.34 &40.29 &43.41 &39.84 &42.90 \\
        \bottomrule
    \end{tabular}
    }
    \caption{Results of transferring attacks against BLIP-2 and InstructBLIP on VQA.}
    \label{tab:transferbility_of_lvlms_vqa}
\end{table*}

\begin{table*}[ht]
    \centering
    \resizebox{\linewidth}{!}{
    \begin{tabular}{c|c|c|c|c|c|c|c|c|c|c}
        \toprule
         LLM of & Attack & \multirow{2}{*}{Loss $\mathcal{L}$} & \multicolumn{8}{c}{Image Caption evaluation on InstructBLIP}  \\
         InstructBLIP & method & & CIDEr & BLEU-1 & BLEU-2 & BLEU-3 & BLEU-4 & ROUGE-L & METEOR & SPICE \\
         \midrule
         \multirow{5.88}{*}{Vicuna-7B} & \cellcolor{gray!30}\ding{56} & \cellcolor{gray!30}\ding{56} & \cellcolor{gray!30}160.5 & \cellcolor{gray!30}82.5 & \cellcolor{gray!30}68.4 & \cellcolor{gray!30}54.6 & \cellcolor{gray!30}42.9 & \cellcolor{gray!30}61.4 & \cellcolor{gray!30}31.4 & \cellcolor{gray!30}25.4 \\
         \cmidrule{2-11}
         & \multirow{2}{*}{PGD} & $\mathcal{L}_\text{LLM}(\text{RSQ}_{10})$ & 71.4 & 61.8 & 42.9 & 29.4 & 20.4 & 45.8 & 19.6 & 13.1 \\
         &  & $\mathcal{L}_\text{QAVA}(\text{RSQ}_{10})$ & 16.9 & 38.8 & 19.8 & 10.4 & 5.9 & 28.2 & 10.0 & 4.1 \\
         \cmidrule{2-11}
         & \multirow{2}{*}{CW} & $\mathcal{L}_\text{LLM}(\text{RSQ}_{10})$ & 92.7 & 67.3 & 50.0 & 36.3 & 25.9 & 49.6 & 22.5 & 16.1 \\
         &  & $\mathcal{L}_\text{QAVA}(\text{RSQ}_{10})$ & 13.0 & 29.2 & 14.5 & 7.1 & 3.8 & 25.6 & 8.6 & 3.4 \\
        \bottomrule
    \end{tabular}
    }
    \caption{Results of inter-task transferability (VQA $\to$ caption) of adversarial examples on InstructBLIP. The grey lines are the results of clean images. $\text{QAVA}(\text{RSQ}_\text{10})$ effectively improves the tasks transferability.}
    \label{tab:caption_iblip}
\end{table*}

\begin{table*}[!ht]
    \centering
    \resizebox{\linewidth}{!}{
    \begin{tabular}{c|c|c|c|c|c|c|c|c|c}
        \toprule
        Target model & Surrogate model & CIDEr &BLEU-1 &BLEU-2 &BLEU-3 &BLEU-4 &ROUGE-L &METEOR &SPICE \\
        \midrule
        \multirow{5}{*}{MiniGPT-4} & \cellcolor{gray!30}Clean image & \cellcolor{gray!30}89.7 & \cellcolor{gray!30}65.9 & \cellcolor{gray!30}49.9 & \cellcolor{gray!30}35.8 & \cellcolor{gray!30}25.0 & \cellcolor{gray!30}52.9 & \cellcolor{gray!30}29.4 & \cellcolor{gray!30}24.6 \\
        & InstructBLIP $\text{FlanT5}_\text{XL}$  &16.9 &39.9 &24.0 &14.0 &8.4 &33.3 &15.9 &8.6 \\
        & InstructBLIP Vicuna-7B &12.9 &37.8 &21.7 &12.3 &7.4 &31.8 &14.8 &7.4 \\
        & InstructBLIP $\text{FlanT5}_\text{XXL}$ &18.0 &39.6 &24.0 &14.5 &9.0 &33.7 &15.9 &8.6 \\
        & InstructBLIP Vicuna-13B &13.7 &36.2 &21.0 &12.0 &7.2 &31.2 &14.1 &6.9 \\
        \midrule
        \multirow{5}{*}{LLaVA} & \cellcolor{gray!30}Clean image & \cellcolor{gray!30}116.9 & \cellcolor{gray!30}73.1 & \cellcolor{gray!30}56.8 & \cellcolor{gray!30}42.0 & \cellcolor{gray!30}30.3 & \cellcolor{gray!30}56.6 & \cellcolor{gray!30}29.9 & \cellcolor{gray!30}24.4 \\
        & InstructBLIP $\text{FlanT5}_\text{XL}$ &82.9 &65.5 &47.8 &33.5 &22.7 &50.2 &25.1 &18.4 \\
        & InstructBLIP Vicuna-7B &82.3 &64.6 &47.4 &33.5 &23.2 &49.7 &25.0 &18.1 \\
        & InstructBLIP $\text{FlanT5}_\text{XXL}$ &86.5 &65.6 &48.1 &33.9 &23.5 &50.5 &25.5 &18.8 \\
        & InstructBLIP Vicuna-13B &79.9 &64.2 &46.5 &32.5 &22.3 &49.2 &24.7 &17.8 \\
        \bottomrule
    \end{tabular}
    }
    \caption{Results of transferability of adversarial examples on both tasks (VQA $\to$ caption) and models (InstructBLIP $\to$ MiniGPT-4/LLaVA). The attack is all $\mathcal{L}_\text{QAVA}(\text{RSQ}_{10})$.}
    \label{tab:caption_llava_minigpt}
\end{table*}

\noindent
\textbf{Transferability of QAVA between InstructBLIP and LLaVA on VQA tasks.} We investigated the transferability of the QAVA attack on the VQA tasks between InstructBLIP and LLaVA, with the results presented in \cref{tab:transfer_vqa_to_llava}. To enhance the transferability of the adversarial examples, we incorporated the momentum attack \cite{dong2018boosting} and diverse input methods \cite{xie2019improving}.

\noindent
\textbf{Transferability of QAVA between BLIP-2 and InstructBLIP on VQA tasks.} We examined the transferability of QAVA across the LVLMs utilized in \cref{tab:more_lvlms_qava}. Adversarial examples were generated on each LVLM independently, and their VQA scores were evaluated on the other models, with the results presented in \cref{tab:transferbility_of_lvlms_vqa}. The experiments were conducted using the VQA v2 32+50 dataset with $\mathcal{L}_\text{QAVA}$($\text{RSQ}_{10}$). Each row represents the surrogate model used to generate the adversarial perturbation, while each column represents the target model used to test the VQA scores. The diagonal cells indicate white-box model settings, whereas the non-diagonal cells represent black-box model settings, demonstrating transferability. The results indicate that our QAVA exhibits strong transferability.

\noindent
\textbf{Transferability of QAVA from VQA to caption task on InstructBLIP.} We investigated the performance of images subjected to adversarial attacks on tasks beyond VQA. Specifically, we input adversarial images generated from the VQA v2 500+10 dataset into InstructBLIP to produce image captions using the prompt ``A short image caption:''. The results of these adversarial images on the image caption task are displayed in \cref{tab:caption_iblip}. Despite targeting unrelated random questions, the adversarial approach effectively reduces image caption performance. The efficacy of QAVA-generated adversarial images on the caption task underscores the effectiveness of RSQ and $\mathcal{L}_\text{QAVA}$. The transferability results for the captioning task on more LVLMs are shown in \cref{tab:caption_iblip_extended} in the Appendix. 

\noindent
\textbf{Transferability of QAVA from InstructBLIP and VQA to LLaVA/MiniGPT-4 and caption task.} We also explored the transferability of QAVA across tasks (from VQA to captioning) and models (from InstructBLIP to LLaVA/MiniGPT-4). The results presented in \cref{tab:caption_llava_minigpt} demonstrate QAVA's strong transferability between tasks and LVLMs, even when the attacks are based on irrelevant random questions. Given that image captioning is a classical pre-training task for LVLMs, employing the adversarial examples generated by QAVA to disrupt the training process of LVLMs could potentially have a significant impact.

\subsection{Generalizability of QAVA}

Our primary experiments were conducted using the VQA v2 dataset. To further assess the generalizability of QAVA, we performed additional experiments on several other datasets, including ImageNet \cite{deng2009imagenet}, OKVQA \cite{marino2019ok}, NoCaps \cite{agrawal2019nocaps}, and Flickr30k \cite{young2014image}. The results presented in \cref{tab:qava_other_dataset} demonstrate the effectiveness of QAVA in executing successful attacks across a diverse array of datasets.

\begin{table}[ht]
    \centering
    \begin{tabular}{c|c|c|c}
        \toprule
        Dataset & Metric & Clean & QAVA \\
        \midrule
        ImageNet & Accuracy & 81.0 & 34.8 \\
        OKVQA & VQA score & 56.9 & 23.85 \\
        NoCaps & CIDEr & 120.2 & 15.2 \\
        Flickr30k & CIDEr & 85.2 & 9.1 \\
        \bottomrule
    \end{tabular}
    \caption{The generalizability of QAVA on other datasets including classification, Q\&A and captioning.}
    \label{tab:qava_other_dataset}
\end{table}

\subsection{Discussions of QAVA}
\label{sec:discussion}
\noindent
\textbf{Potential Further Optimizations of QAVA}. (1) Recent research on universal adversarial attacks on images has introduced Stochastic Gradient Aggregation (SGA) \cite{liu2023enhancing}, a technique that improves stability by calculating multiple gradients over a small batch of images and merging them into a single gradient. 

Inspired by SGA, we can further improve the attack performance of QAVA by sampling different small batches of stochastic questions at each step of the attack process, as illustrated in \cref{tab:qava_sga}. (2) Further optimization could enhance the performance of QAVA. Specifically, the vulnerable vision-language alignment module consists of multiple layers, and our current loss function $\mathcal{L}_{QAVA}$ targets only the output of the last layer of the Q-former. We extended the loss function to incorporate outputs from all layers of the Q-former. The results presented in \cref{tab:qava_multi_layer} demonstrate that this minor optimization of the loss function leads to an improvement in the performance of QAVA.

\begin{table}[ht]
    \centering
    \resizebox{\linewidth}{!}{
    \begin{tabular}{c|c|c|c|c}
        \toprule
        \multirow{2}{*}{Attack method} & \multicolumn{4}{c}{VQA v2 scores} \\
         & Overall & Other & Number & Yes/No \\
        \midrule
        $\mathcal{L}_\text{QAVA}$($\text{RSQ}_{10}$) & 44.85 & 22.11 & 43.25 & 73.83 \\
        $\mathcal{L}_\text{QAVA}$($\text{RSQ}_{10}$) + SGA & 41.14 & 19.08 & 37.97 & 69.74 \\
        \bottomrule
    \end{tabular}
    }
    \caption{The optimized version of QAVA, drawing inspiration from SGA, leads to more robust query-agnostic adversarial examples.}
    \label{tab:qava_sga}
\end{table}

\begin{table}[ht]
    \centering
    \begin{tabular}{c|c}
        \toprule
        Method & VQA v2 scores \\
        \midrule
        $\mathcal{L}_\text{QAVA}$($\text{RSQ}_{25}$) & $44.07\pm0.83$ \\
        Multi-layer $\mathcal{L}_\text{QAVA}$($\text{RSQ}_{25}$) & $42.54\pm0.92$ \\
        \bottomrule
    \end{tabular}
    \caption{The optimized version of QAVA, employing multi-layer loss function $\mathcal{L}_\text{QAVA}$.}
    \label{tab:qava_multi_layer}
\end{table}

\noindent
\textbf{Potential Defense of QAVA}. We discuss possible defenses against QAVA in \cref{sec:defense}.

\section{Conclusion}
\label{sec:conclusion}

\noindent
In this paper, we introduce a robust adversarial attack method, QAVA, designed to generate adversarial examples that significantly mislead responses to unknown target questions for a given image. QAVA initially identifies the vulnerability within the LVLMs, specifically the vision-language alignment module. Subsequently, it executes adversarial attacks utilizing a broader range of randomly sampled, image-irrelevant questions. Extensive experiments demonstrate the effectiveness of QAVA in both white-box and black-box attack scenarios. Additionally, we verify the high transferability of QAVA across various LVLMs and different tasks. Our findings with QAVA serve as a critical alert regarding the security vulnerabilities of LVLMs.

\section{Limitation}
\label{sec:limitation}

We summarize the limitations of our work as follows. We will try to do these in the future.

(1) Although extensive experiments have demonstrated the effectiveness of attacks utilizing irrelevant questions, we have not yet provided a plausible explanation for the impact that such irrelevant randomized questions can have on the attacks.

(2) Insufficient assessment of potential negative impacts of QAVA. As we analyzed, the adversarial examples obtained using QAVA are more aggressive and may have a larger negative impact if they are used for poisoning in the pre-training or supervised fine-tuning process of LVLMs. However, this aspect was not evaluated experimentally.

(3) We only evaluated the transferring attack of QAVA for image captioning tasks, not for broader visual-language tasks.

\section{Acknowledgments}
This work was supported by the National Natural Science Foundation of China (62376024, 62325405), the Young Elite Scientists Sponsorship Program by CAST (2023QNRC001) and Beijing National Research Center for Information Science and Technology (BNRist, BNR2024TD03001).

\clearpage
\bibliography{custom}

\clearpage
\appendix

\section{The Algorithm of QAVA}
\label{sec:algorithm}
\begin{algorithm}[ht]
    \SetKwInput{KwModel}{Model}
    \SetKwInput{KwHyperparameter}{Hyperparameter}
    \caption{The steps of $\text{QAVA}(\text{RSQ}_\text{N})$ using PGD to attack Q-former}
    \label{alg:example}
    \KwIn{image $x_i$}
    \KwModel{visual encoder $f_i$, Q-former $Q$}
    \KwData{the question set $\mathcal{T}$}
    \KwHyperparameter{questions number $N$, attack step $\alpha$, perturbation limitation $\epsilon_\infty$, number of attack iterations $n$}
    \KwOut{the adversarial image $x'_i$}
        \For{$k\gets 1$ \KwTo $n$}
        {
            $\mathcal{L}_\text{QAVA}=0$\;
            \For{$j\gets 1$ \KwTo $N$}
            {
                \tcc{Randomly select a question from $\mathcal{T}$}
                $x_t\sim\mathcal{T}$\;
                $q=\mathcal{Q}(f_i(x_i), x_t)$\;
                $q'=\mathcal{Q}(f_i(x'_i), x_t)$\;
                $\mathcal{L}_\text{QAVA}=\mathcal{L}_\text{QAVA}+\frac{1}{N}\mathcal{L}_\text{QAVA}(q, q')$\;
            }
             $x'_i=\text{Clip}_{\epsilon_\infty}(x'_i+\alpha\times\text{sign}(\nabla_{x'_i}\mathcal{L}_\text{QAVA}))$\;
        }
    \Return{$x'_i$}\;
\end{algorithm}

\section{The Discussion of Potential Defense Methods for QAVA}
\label{sec:defense}

QAVA targets the output of the vulnerable visual-language alignment module within LVLMs. To mitigate such attacks, potential defense strategies include: (1) Implementing adversarial training of the visual-language alignment module. This approach is cost-effective, as it does not involve the computationally intensive LLM. (2) Developing mechanisms to suppress the module's output when faced with image-unrelated questions. For instance, earlier LVLMs \cite{qi2023visual} were susceptible to jailbreak attacks via adversarial images irrelevant to the input instructions. Conversely, modern LVLMs, such as GPT-4 and Gemini, would ignore input images unrelated to the instructions. However, these models might still be vulnerable to jailbreaks when adversarial images pertain to the instructions.

\section{Visualization results on the QAVA attack}

\begin{figure*}[ht]
    \centering
    \begin{subfigure}{0.28\linewidth}
        \includegraphics[width=\linewidth]{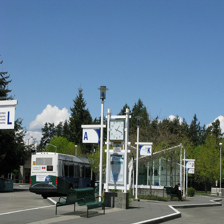}
    \end{subfigure}
    \begin{subfigure}{0.28\linewidth}
        \includegraphics[width=\linewidth]{figure-supp/COCO_val2014_000000038540_adv.png}
    \end{subfigure}
    \begin{subfigure}{0.28\linewidth}
        \includegraphics[width=\linewidth]{figure-supp/COCO_val2014_000000038540_adv_ssah.png}
    \end{subfigure}
    \hfill
    \begin{subfigure}{0.28\linewidth}
        \includegraphics[width=\linewidth]{figure-supp/COCO_val2014_000000079837_ori.png}
    \end{subfigure}
    \begin{subfigure}{0.28\linewidth}
        \includegraphics[width=\linewidth]{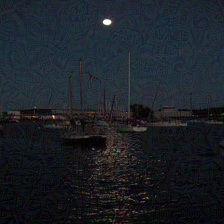}
    \end{subfigure}
    \begin{subfigure}{0.28\linewidth}
        \includegraphics[width=\linewidth]{figure-supp/COCO_val2014_000000079837_adv_ssah.png}
    \end{subfigure}
    \hfill
    \begin{subfigure}{0.28\linewidth}
        \includegraphics[width=\linewidth]{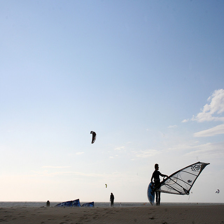}
    \end{subfigure}
    \begin{subfigure}{0.28\linewidth}
        \includegraphics[width=\linewidth]{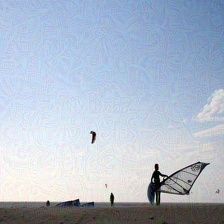}
    \end{subfigure}
    \begin{subfigure}{0.28\linewidth}
        \includegraphics[width=\linewidth]{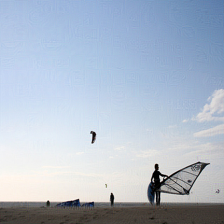}
    \end{subfigure}
    \hfill
    \begin{subfigure}{0.28\linewidth}
        \includegraphics[width=\linewidth]{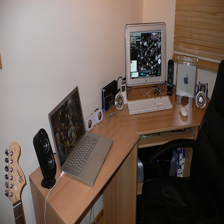}
    \end{subfigure}
    \begin{subfigure}{0.28\linewidth}
        \includegraphics[width=\linewidth]{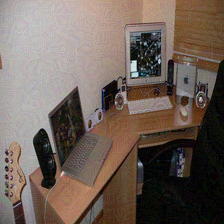}
    \end{subfigure}
    \begin{subfigure}{0.28\linewidth}
        \includegraphics[width=\linewidth]{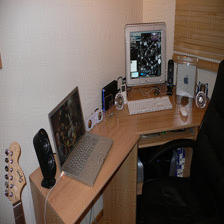}
    \end{subfigure}
    \begin{subfigure}{0.28\linewidth}
        \includegraphics[width=\linewidth]{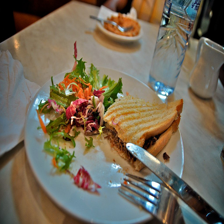}
        \caption{Clean image with VQA score 78.00.}
        \label{fig:imperceptibility_clean}
    \end{subfigure}
    \begin{subfigure}{0.28\linewidth}
        \includegraphics[width=\linewidth]{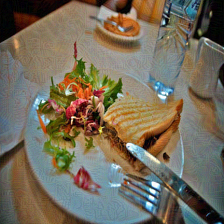}
        \caption{QAVA adversarial image with VQA score: 44.85.}
        \label{fig:imperceptibility_qava}
    \end{subfigure}
    \begin{subfigure}{0.28\linewidth}
        \includegraphics[width=\linewidth]{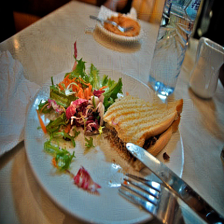}
        \caption{QAVA+SSAH adversarial image with VQA score: 50.67.}
        \label{fig:imperceptibility_ssah}
    \end{subfigure}
    \caption{The clean images, the QAVA adversarial images, and the QAVA+SSAH adversarial images. All experiments are conducted using InstructBLIP Vicuna-7B with the attack $\mathcal{L}_\text{QAVA}$($\text{RSQ}_\text{10}$).}
    \label{fig:figure_imperceptibility}
\end{figure*}

In \cref{sec:abalation_imperceptibility}, we examine the imperceptibility of adversarial examples produced by the QAVA attack. The visual representations of these adversarial examples are provided in \cref{fig:figure_imperceptibility}.

\section{Results of the extension to more LVLMs of \cref{{tab:caption_iblip}}}
\Cref{tab:caption_iblip} presents the inter-task transferability of QAVA on InstructBLIP Vicuna-7B. Additionally, \cref{tab:caption_iblip_extended} provides further results for other versions of InstructBLIP, demonstrating QAVA's applicability across different LVLMs.

\section{More results on $\text{FlanT5}_\text{XL}$}
\label{sec:more_results_on_flant5xl}

For \Cref{tab:rsq_loss_QAVA,tab:VQG} in the main paper, we also conducted the same experiments on $\text{FlanT5}_\text{XL}$. The results are shown in \cref{tab:rsq_loss_QAVA_flan,tab:VQG_flan}.

\begin{table*}[ht]
    \centering
    \resizebox{\linewidth}{!}{
    \begin{tabular}{c|c|c|c|c|c|c|c|c|c|c}
        \toprule
         LLM of & Attack & \multirow{2}{*}{Loss $\mathcal{L}$} & \multicolumn{8}{c}{Image Caption evaluation on InstructBLIP}  \\
         InstructBLIP & method & & CIDEr & BLEU-1 & BLEU-2 & BLEU-3 & BLEU-4 & ROUGE-L & METEOR & SPICE \\
         \midrule
         \multirow{5.88}{*}{$\text{FlanT5}_\text{XL}$} & \cellcolor{gray!30}\ding{56} & \cellcolor{gray!30}\ding{56} & \cellcolor{gray!30}154.6 & \cellcolor{gray!30}81.9 & \cellcolor{gray!30}67.1 & \cellcolor{gray!30}52.0 & \cellcolor{gray!30}39.0 & \cellcolor{gray!30}59.7 & \cellcolor{gray!30}30.2 & \cellcolor{gray!30}24.5 \\
         \cmidrule{2-11}
         & \multirow{2}{*}{PGD} & $\mathcal{L}_\text{LLM}(\text{RSQ}_{10})$ & 52.4 & 57.1 & 37.7 & 24.2 & 15.6 & 42.2 & 17.0 & 10.0 \\
         &  & $\mathcal{L}_\text{QAVA}(\text{RSQ}_{10})$ & 19.7 & 37.8 & 20.6 & 11.8 & 7.1 & 28.4 & 10.5 & 4.3 \\
         \cmidrule{2-11}
         & \multirow{2}{*}{CW} & $\mathcal{L}_\text{LLM}(\text{RSQ}_{10})$ & 87.5 & 66.5 & 48.7 & 34.8 & 24.3 & 48.0 & 21.6 & 15.1 \\
         &  & $\mathcal{L}_\text{QAVA}(\text{RSQ}_{10})$ & 6.8 & 25.3 & 11.6 & 4.8 & 2.3 & 24.3 & 7.7 & 2.4 \\
        \midrule
         \multirow{5.88}{*}{$\text{FlanT5}_\text{XXL}$} & \cellcolor{gray!30}\ding{56} & \cellcolor{gray!30}\ding{56} & \cellcolor{gray!30}154.0 & \cellcolor{gray!30}82.3 & \cellcolor{gray!30}67.4 & \cellcolor{gray!30}52.9 & \cellcolor{gray!30}40.6 & \cellcolor{gray!30}60.9 & \cellcolor{gray!30}30.3 & \cellcolor{gray!30}24.3 \\
         \cmidrule{2-11}
         & \multirow{2}{*}{PGD} & $\mathcal{L}_\text{LLM}(\text{RSQ}_{10})$ & 43.1 & 54.6 & 34.8 & 22.0 & 14.4 & 40.6 & 16.1 & 8.8 \\
         &  & $\mathcal{L}_\text{QAVA}(\text{RSQ}_{10})$ & 23.2 & 41.4 & 23.2 & 12.3 & 6.7 & 30.3 & 11.3 & 5.6 \\
         \cmidrule{2-11}
         & \multirow{2}{*}{CW} & $\mathcal{L}_\text{LLM}(\text{RSQ}_{10})$ & 88.8 & 58.6 & 42.8 & 30.1 & 21.3 & 47.7 & 20.7 & 15.2 \\
         &  & $\mathcal{L}_\text{QAVA}(\text{RSQ}_{10})$ & 11.6 & 23.8 & 11.6 & 5.7 & 3.0 & 25.2 & 7.9 & 3.1 \\
        \midrule
         \multirow{5.88}{*}{Vicuna-13B} & \cellcolor{gray!30}\ding{56} & \cellcolor{gray!30}\ding{56} & \cellcolor{gray!30}129.2 & \cellcolor{gray!30}62.6 & \cellcolor{gray!30}50.1 & \cellcolor{gray!30}38.6 & \cellcolor{gray!30}28.8 & \cellcolor{gray!30}57.0 & \cellcolor{gray!30}28.8 & \cellcolor{gray!30}23.6 \\
         \cmidrule{2-11}
         & \multirow{2}{*}{PGD} & $\mathcal{L}_\text{LLM}(\text{RSQ}_{10})$ & 60.6 & 49.0 & 34.1 & 22.9 & 15.2 & 42.7 & 19.1 & 12.7 \\
         &  & $\mathcal{L}_\text{QAVA}(\text{RSQ}_{10})$ & 12.7 & 25.4 & 12.6 & 6.0 & 3.2 & 25.9 & 8.9 & 3.6 \\
         \cmidrule{2-11}
         & \multirow{2}{*}{CW} & $\mathcal{L}_\text{LLM}(\text{RSQ}_{10})$ & 72.5 & 52.0 & 36.7 & 25.4 & 17.4 & 44.6 & 20.9 & 14.8 \\
         &  & $\mathcal{L}_\text{QAVA}(\text{RSQ}_{10})$ & 7.2 & 25.0 & 11.7 & 5.4 & 2.9 & 22.4 & 7.8 & 3.0 \\
        \bottomrule
    \end{tabular}
    }
    \caption{Results of inter-task transferability (VQA $\to$ caption) of adversarial examples on InstructBLIP. The grey lines are the results of clean images. $\text{QAVA}(\text{RSQ}_\text{10})$ effectively improves the tasks transferability.}
    \label{tab:caption_iblip_extended}
\end{table*}

\begin{table}[ht]
    \centering
    \resizebox{\linewidth}{!}{
    \begin{tabular}{c|c|c|c|c|c|c}
        \toprule
        Attack & Loss & Question & \multicolumn{4}{c}{InstructBLIP($\text{FlanT5}_\text{XL}$)} \\
        method & $\mathcal{L}$ & strategy & Overall & Other & Number & Yes/No \\
        \midrule
        \rowcolor{gray!30}\ding{56} & \ding{56} & \ding{56} & 74.54 & 63.07 & 66.52 & 91.31 \\ 
        \midrule
        \multirow{9.44}{*}{PGD} & \multirow{2}{*}{$\mathcal{L}_\text{LLM}$} & $\text{WTQ}_\text{50}$ & 40.91 & 18.77  & 31.02 & 71.63 \\
        & & $\text{RSQ}_\text{25}$ & 49.78{\scriptsize ($\pm$0.92)} & 26.98  & 46.60 & 79.33 \\
        \cmidrule{2-7}
        & \multirow{7}{*}{$\mathcal{L}_\text{QAVA}$} & $\text{WTQ}_\text{50}$ & 36.11 & 17.71  & 34.33 & 59.70 \\
        & & $\text{RSQ}_\text{1}$ & 41.54{\scriptsize ($\pm$1.55)} & 22.94  & 41.50 & 64.87 \\
        & & $\text{RSQ}_\text{5}$ & 36.30{\scriptsize ($\pm$1.89)} & 19.76  & 32.81 & 58.07 \\
        & & $\text{RSQ}_\text{10}$ & 34.31{\scriptsize ($\pm$2.06)} & 18.10  & 31.61 & 55.42 \\
        & & $\text{RSQ}_\text{15}$ & 35.22{\scriptsize ($\pm$1.28)} & 17.76  & 33.13 & 57.72 \\
        & & $\text{RSQ}_\text{20}$ & 35.50{\scriptsize ($\pm$1.21)} & 18.14  & 34.50 & 57.56 \\
        & & $\text{RSQ}_\text{25}$ & 35.89{\scriptsize ($\pm$1.70)} & 18.47  & 35.19 & 57.92 \\
        \midrule
        \multirow{9.44}{*}{CW} & \multirow{2}{*}{$\mathcal{L}_\text{LLM}$} & $\text{WTQ}_\text{50}$ & 41.95 & 20.08  & 29.09 & 73.21 \\
        & & $\text{RSQ}_\text{25}$ & 52.77{\scriptsize ($\pm$0.90)} & 32.04  & 47.13 & 80.45 \\
        \cmidrule{2-7}
        & \multirow{7}{*}{$\mathcal{L}_\text{QAVA}$} & $\text{WTQ}_\text{50}$ & 9.61 & 4.83  & 5.29 & 16.87 \\
        & & $\text{RSQ}_\text{1}$ & 20.47{\scriptsize ($\pm$4.21)} & 10.42  & 21.25 & 32.83 \\
        & & $\text{RSQ}_\text{5}$ & 11.99{\scriptsize ($\pm$1.99)} & 5.43  & 11.31 & 20.42 \\
        & & $\text{RSQ}_\text{10}$ & 10.62{\scriptsize ($\pm$1.55)} & 5.08  & 7.79 & 18.41 \\
        & & $\text{RSQ}_\text{15}$ & 12.01{\scriptsize ($\pm$1.42)} & 5.59  & 9.99 & 20.68 \\
        & & $\text{RSQ}_\text{20}$ & 11.46{\scriptsize ($\pm$1.92)} & 5.40  & 9.89 & 19.52 \\
        & & $\text{RSQ}_\text{25}$ & 10.67{\scriptsize ($\pm$2.29)} & 4.64  & 8.92 & 18.75 \\
        \bottomrule
    \end{tabular}
    }
    \caption{$\mathcal{L}_\text{QAVA}$ is better than $\mathcal{L}_\text{LLM}$ in use of $\text{RSQ}_\text{N}$.}
    \label{tab:rsq_loss_QAVA_flan}
\end{table}

\begin{table}[ht]
    \centering
    \resizebox{\linewidth}{!}{
    \begin{tabular}{c|c|c|c|c|c}
        \toprule
         Attack & Question & \multicolumn{4}{c}{InstructBLIP($\text{FlanT5}_\text{XL}$)} \\
         method & strategy & Overall & Other & Number & Yes/No \\
        \midrule
         \multirow{6.76}{*}{PGD} & $\text{WTQ}_\text{50}$ & \cellcolor{gray!30}36.11 & \cellcolor{gray!30}17.71 & \cellcolor{gray!30}34.33 & \cellcolor{gray!30}59.70 \\
         \cmidrule{2-6}
         & $\text{RSQ}_\text{10}$ & 34.31{\scriptsize ($\pm$2.06)} & 18.10  & 31.61 & 55.42 \\
         \cmidrule{2-6}
         & $\text{RSQ}^\text{t}_\text{10}$ & 34.54{\scriptsize ($\pm$0.70)} & 18.23  & 29.67 & 56.43 \\
         \cmidrule{2-6}
         & $\text{RSQ}^\text{c}_\text{10}$ & 34.28{\scriptsize ($\pm$2.40)} & 18.36  & 31.93 & 54.94 \\
         \cmidrule{2-6}
         & $\text{VQG}_\text{10}$ & 35.83{\scriptsize ($\pm$2.48)} & 18.81  & 35.18 & 57.36 \\
        \midrule
         \multirow{6.76}{*}{CW} & $\text{WTQ}_\text{50}$ & \cellcolor{gray!30}9.61 & \cellcolor{gray!30}4.83  & \cellcolor{gray!30}5.29 & \cellcolor{gray!30}16.87 \\
         \cmidrule{2-6}
         & $\text{RSQ}_\text{10}$ & 10.62{\scriptsize ($\pm$1.55)} & 5.08  & 7.79 & 18.41 \\
         \cmidrule{2-6}
         & $\text{RSQ}^\text{t}_\text{10}$ & 9.87{\scriptsize ($\pm$1.01)} & 4.94  & 8.66 & 16.41 \\
         \cmidrule{2-6}
         & $\text{RSQ}^\text{c}_\text{10}$ & 9.76{\scriptsize ($\pm$2.89)} & 4.83  & 7.24 & 16.69 \\
         \cmidrule{2-6}
         & $\text{VQG}_\text{10}$ & 9.04{\scriptsize ($\pm$1.84)} & 4.61  & 7.58 & 15.04 \\
        \bottomrule
    \end{tabular}
    }
    \caption{The ablation study of the question sampling strategy as outlined in \cref{sec:rsq}.}
    \label{tab:VQG_flan}
\end{table}

\end{document}